\documentclass{article}
\usepackage{spconf}
\ninept
\usepackage{microtype}
\usepackage{graphicx}
\usepackage{subcaption}
\usepackage{array,booktabs} 
\usepackage[T1]{fontenc}
\usepackage{amsmath}
\usepackage{mathtools}
\usepackage{multirow}
\usepackage{multicol}
\usepackage{enumitem}
\usepackage{url}
\usepackage{xcolor}

\title{Do as I mean, not as I say: \\ Sequence Loss Training for Spoken Language Understanding}
\name{\parbox{\linewidth}{\centering Milind Rao, Pranav Dheram, Gautam Tiwari, Anirudh Raju, \\ Jasha Droppo, Ariya Rastrow, Andreas Stolcke }\thanks{© 20XX IEEE. Personal use of this material is permitted. Permission from IEEE must be obtained for all other uses, in any current or future media, including reprinting/republishing this material for advertising or promotional purposes, creating new collective works, for resale or redistribution to servers or lists, or reuse of any copyrighted component of this work in other works.}}
\address{
   Amazon Alexa, USA, \\
  {\normalsize \ttfamily \{milinrao,pddheram,tgautam,ranirudh,drojasha,arastrow,stolcke\}@amazon.com}
}

\def \figpath {./}

\begin{document}
\setlength{\abovedisplayskip}{2pt}
\setlength{\belowdisplayskip}{0pt}
\setlength\tabcolsep{5 pt}

\maketitle

\begin{abstract}
Spoken language understanding (SLU) systems extract transcriptions, as well as semantics of intent or named entities from speech, and are essential components of voice activated systems. SLU models, which either directly extract semantics from audio or are composed of pipelined automatic speech recognition (ASR) and  natural language understanding (NLU) models, are typically trained via differentiable cross-entropy losses, even when the relevant performance metrics of interest are word or semantic error rates. In this work, we propose non-differentiable sequence losses based on SLU metrics as a proxy for semantic error and use the REINFORCE trick to train ASR and SLU models with this loss. We show that custom sequence loss training is the state-of-the-art on open SLU datasets and leads to 6\% relative improvement in both ASR and NLU performance metrics on large proprietary datasets. We also demonstrate how the semantic sequence loss training paradigm can be used to update ASR and SLU models without transcripts, using semantic feedback alone.  

\end{abstract}


\vspace{-0.4cm}
\section{Introduction}
\label{sec:introduction}
\vspace{-0.3cm}

Spoken language understanding systems that aim to understand user commands are an integral part of voice interfaces or spoken dialogue systems. Our focus is on developing compact models that can be deployed on edge devices allowing low-latency processing without transmitting audio and/or transcripts to cloud servers and enabling offline use in remote, medical, vehicular, or emergency environments. Table \ref{tab:utt_eg} shows an example of the transcript and semantics of an utterance. A conventional deployment for SLU comprises two distinct pipelined stages: (1) ASR to transcribe utterances (2) an NLU system that consumes the transcription and produces utterance intent and named entities or slots. 

\vspace{-0.4cm}
\subsection{Prior Work}
\vspace{-0.2cm}

A pipelined or compositional deployment would make use of end-to-end (E2E) ASR architectures such as RNN-T  \cite{graves2012sequence}, CTC \cite{graves2006connectionist}, Transformer-transducers \cite{karita2019comparative}, LAS \cite{chan16}, or conventional RNN-HMM hybrid ASR systems \cite{chiu2018state}.  Extracting intent and slots from transcripts is a long running problem in NLU \cite{lample2016neural, kim2017onenet, wu2017clinical} that uses LSTMs or Transformers \cite{chen2019bert,bunk2020diet}. The interface between ASR and NLU systems has traditionally been the single best hypothesis generated by ASR, although richer interfaces such as lattices and word confusion networks have also been proposed \cite{hakkani2006beyond, henderson2012discriminative, tur2002improving, huang2019adapting}.  

With the compositional approach listed above, ASR errors cascade down to the NLU system, ASR is not trained aware of downstream NLU use, and NLU is not trained to compensate for ASR ambiguity or error. \cite{haghani2018audio} first introduced multi-stage, multi-task and joint models for E2E SLU. Most prior work in this space \cite{ghannay2018end, serdyuk2018towards, qian2017exploring, tomashenko2020dialogue, lugosch2019speech} directly computes a serialization of the semantics without intermediate text output. Another common approach uses transfer learning of pretrained ASR models to SLU tasks by replacing the final layer. In contrast, \cite{Rao2020Speech} used pretrained ASR models and NLU architectures and replaced the one-best ASR hypothesis interface with a neural network interface allowing joint training of ASR and NLU.

ASR systems are typically first trained with differentiable losses such as cross-entropy (CE), CTC or RNN-T. NLU systems are trained using CE losses for classification problems like intent, domain, or named entity tags. E2E SLU systems make use of cross-entropy on either transcripts, intents, slots, or some serialization of semantics. The CE metric is simply a proxy for and does not directly minimize SLU metrics of interest. REINFORCE  \cite{williams1992simple} can be used to train with arbitrary non-differentiable loss functions. This was extended to  mWER training for ASR \cite{juang1997minimum}, LAS \cite{prabhavalkar2018minimum}, and RNN-T \cite{guo2020efficient}. REINFORCE corresponds to the policy gradient approach among other reinforcement learning methods for seq2seq networks \cite{keneshloo2019deep}.

\begin{table}
\caption{An example of intent, slots for an utterance. \label{tab:utt_eg}}
\vspace{-0.3cm}
\begin{tabular}{| p{0.2\linewidth} | p{0.7\linewidth} |}
\hline
Transcript& set an alarm for six a.m \\ \hline 
Intent & SetNotificationIntent \\ \hline 
Slots & NotificationType - alarm, Time - six a.m. \\ \hline
\end{tabular}
\vspace{-0.7cm}
\end{table}

\vspace{-0.5cm}
\subsection{Contributions}
\vspace{-0.2cm}

\begin{figure*}
    \centering
    \includegraphics[width=0.8\linewidth]{\figpath 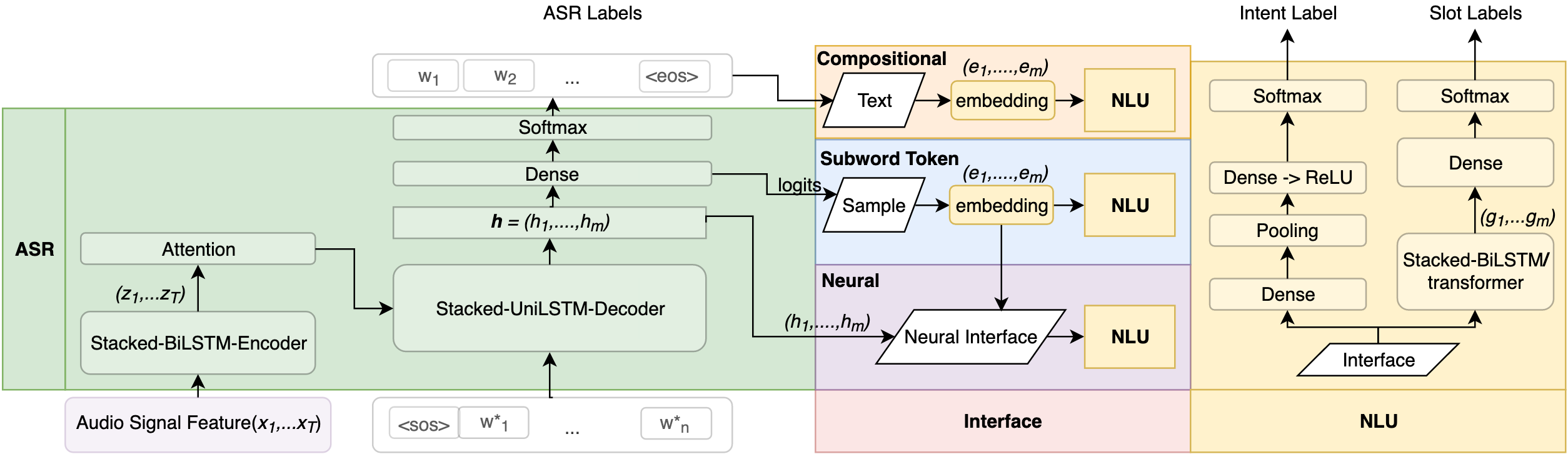}
    \caption{E2E SLU architectures including ASR subsystem, neural NLU subsystem and 3 interfaces - token, text and neural \label{fig:e2e_slu}}
    \includegraphics[width=0.9\linewidth]{\figpath 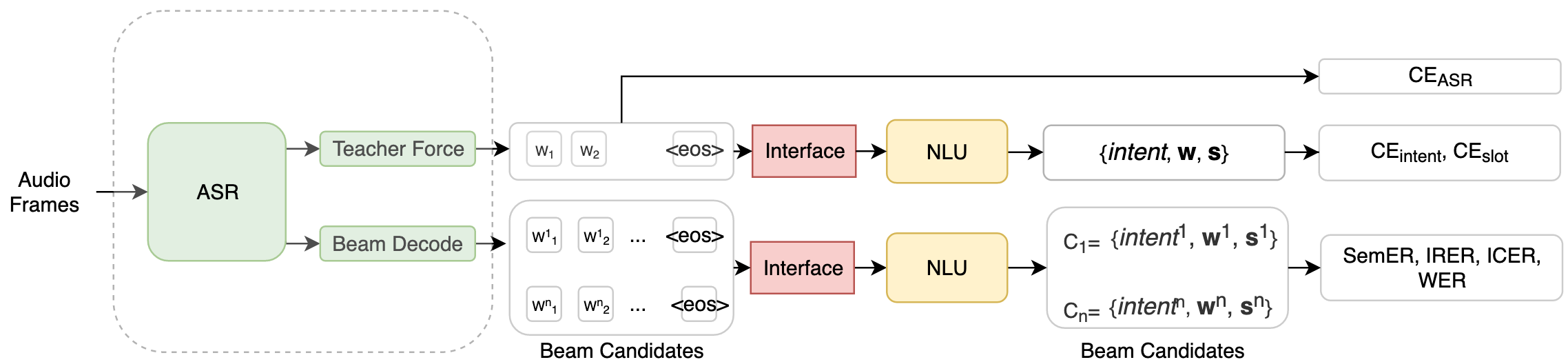}
    \caption{Training SLU models with non-differentiable sequence losses. Dotted box encompasses ASR model including decoder \label{fig:seq_loss}}
    \vspace{-0.5cm}
\end{figure*}

We consider the class of SLU models composed of multistage ASR and NLU subsystems, connected via text, subword tokens, or neural  interfaces that can be \emph{jointly} trained. In these systems, ASR is trained with backpropagation of semantic feedback from NLU, and NLU is trained to be aware of ASR ambiguity and errors preventing a downward cascade of ASR errors as seen in \emph{compositional} systems. We consider ASR systems based on LAS\cite{chan16} and LSTM- or Transformer-encoder-based NLU systems.  

We first develop custom sequence loss training approaches to make use of non-differentiable arbitrary risk values or losses on the entire sequence of outputs. Similar to minimum-word-error-rate (mWER) training, we develop minimum-semantic-error-rate (mSemER) training that directly minimizes intent and slot errors. We introduce alternatives which additionally factor in interpretation (concept) and word errors. Using datasets with complete ASR transcriptions and NLU annotations, we first show significant gains in both ASR and NLU metrics using sequence loss training across datasets ranging from 15 to 15,000 hours for limited to general use cases. We beat all known external benchmarks in the open Fluent speech dataset \cite{lugosch2019speech}.  

As a further application of sequence loss training, we show how a dataset with audio and semantic annotations without human transcriptions can still be used to drive ASR and SLU model improvements.

\vspace{-0.3cm}
\section{Technical Approach}
\label{sec:technical_approach}

\vspace{-0.3cm}
\subsection{ASR-Interface-NLU Models}
\label{sec:models}
\vspace{-0.2cm}

We consider SLU models that comprise an ASR subsystem and an NLU subsystem connected by an interface that passes the 1-best or sampled ASR hypotheses, or via a neural network hidden layer. 

\noindent\textbf{The ASR subsystem} is an attention-based Listen Attend and Spell (LAS) model as shown in the green box of Fig.\ \ref{fig:e2e_slu}. The LAS used here primarily comprises two components - a stacked RNN encoder that encodes audio frames $\mathbf{x}$ to generate representations, and an auto-regressive RNN decoder that sequentially generates logits or subword probability distribution $p_{w, i}(w_i)=\mathbf{P}(w_i|\{w\}_1^{i-1}, \mathbf{x})$ at each decoding step by using multiple attention heads to attend to the audio encoding.

In this work, we focus on the LAS ASR subsystem, but the results can be extended to other architectures, such as streaming-compatible RNN-T systems or Transformer-based ASR architectures.  

\noindent\textbf{The Neural NLU subsystem} as shown in the yellow box of Fig.\ \ref{fig:e2e_slu} accepts a sequence of embeddings or features of tokens decoded by ASR and passes it through multiple BLSTM layers. The outputs of the final layer are used to generate probability distribution $p_{s,i}(s_i)=\mathbf{P}(s_i|\mathbf{w}, \mathbf{x})$ of the slots $\mathbf{s}$ of each subword-token. The slot of a word is the slot of the last token of the word. The outputs or features of the final layer are also max-pooled and passed through feed-forward networks to perform the sequence classification task of obtaining logits over the utterance intent $p_\textrm{intent}(\textrm{intent})=\mathbf{P}(\textrm{intent}|\mathbf{w}, \mathbf{x})$. 

We also present results using a Transformer-encoder NLU architecture. The Transformer-encoder replaces the BLSTM and applies multiple layers of self-attention to embeddings/features of the transcript tokens to produce latent representations that are used to obtain slot and intent logits.  

\noindent\textbf{ASR-NLU Interfaces}. We design the SLU system which comprises multistage ASR and NLU systems with a choice of interfaces:
\begin{itemize}[leftmargin=*]
\item \textbf{Text} from the ASR hypothesis is the interface between ASR and NLU, and embeddings of these tokenized text are the inputs to the NLU system. This allows use of pre-trained ASR and NLU models, using transcribed audio datasets for the former and text-only NLU datasets for the latter. We also term this the \emph{compositional} baseline model that chains pretrained  ASR and NLU.   
\item \textbf{Subword tokens} sampled from the posteriors produced by the ASR decoder form the interface with NLU. ASR and NLU can be jointly trained with NLU trained aware of ASR errors. The Gumbel-softmax sampling approach \cite{jang2016categorical} allows backpropagation of semantic feedback to ASR through the categorical subword token interface. 
\item \textbf{Neural network interface} computes the feature for the token at decoder step $i$ using the token embedding concatenated with the hidden output layer from the LAS decoder LSTM. This interface allows NLU to be trained aware of ASR error, local ASR decoding ambiguity as well as the audio context and allows for ASR to be trained with semantic backpropagation. A pretrained ASR model can be used using transcribed audio. These models are also termed \emph{joint} models in this work.
\end{itemize} 

\vspace{-0.5cm}
\subsection{Loss Functions and Performance Metrics}
\label{sec:metrics}
\vspace{-0.2cm}

Traditionally, differentiable cross-entropy loss functions are used in ASR or SLU model training. The ASR system is \emph{teacher-forced} with the ground truth transcript subword sequence $\mathbf{w}$ and the cross-entropy loss $CE_{\textrm{asr}} = -\sum_{i} \log p_{w, i}(w_i)$ is calculated using the one-step ahead decoded subword probability sequence. NLU consumes the features from ASR and is trained with an intent loss $CE_\textrm{intent} = -\log p_\textrm{intent}(\textrm{intent})$ and a slot-loss $CE_\textrm{slot} = -\sum_i \log p_{s, i}(s_i)$ using ground-truth intent and slot sequence $\mathbf{s}$. 

During joint ASR-NLU model multi-task training, a linear combination of these loss functions is used:
\begin{align}
CE_\textrm{total} &= CE_{\textrm{asr}} + CE_\textrm{intent} + CE_\textrm{slot} \label{eq:x_ent}
\end{align}

While the versatile cross-entropy metric allows for end-to-end model training, it serves merely as a differentiable proxy and does not directly optimize for the final SLU metrics of interest, such as:

\noindent\textbf{Speech recognition}: Word error rate (WER) computed as the ratio of word edit distance (the length of the shortest sequence of insert, delete, and substitute operations over words to transform the hypothesis to the reference) to sequence length. A slot-WER metric that upweights critical words and not carrier phrases may also be used. 

\noindent\textbf{Intent classification}: Intent classification error rate (ICER) is the primary metric for evaluating intent. This is a recall-based metric. 

\noindent\textbf{Slot filling}: Semantic error rate (SemER) metric is used to evaluate jointly the intent and slot-filling performance or NLU performance. Comparing a reference of words and their accompanying tags, performance is classified as: (1) Correct slots - slot name and slot value correctly identified, (2) Deletion errors - slot name present in reference but not hypothesis, (3) Insertion errors - extraneous slot names included by hypothesis, (4) Substitution errors - correct slot name in hypothesis but with incorrect slot value. Intent classification errors are substitution errors.
\begin{align}
\text{SemER} &= \frac{\text{\#Deletion} + \text{\#Insertion} + \text{\#Substitution}}{\underbrace{\text{\#Correct} + \text{\#Deletion} + \text{\#Substitution}}_{\text{\#Slots in Reference}}}
\end{align}
The \textbf{interpretation error rate (IRER)} metric, also known as concept error rate, or simply SLU accuracy is related and is the fraction of utterances for which a semantic error has been made. 

For internal datasets, we report \% relative improvements in these metrics. For example, IRERR is IRER-relative reduction. 

\vspace{-0.5cm}
\subsection{Sequence Loss Training}
\label{sec:sequence_training}
\vspace{-0.2cm}

We make use of the REINFORCE framework \cite{williams1992simple,prabhavalkar2018minimum} to directly optimize for a non-differentiable semantic metric $M(C)$ of interest on random candidate $C = \{\mathbf{w}, \mathbf{s}, \textrm{intent}\}\in\mathcal{C}$ that the SLU model with weight $\theta$ produces with probability $\mathrm{Pr}(C=c|\mathbf{x}) = p(c; \theta) = p_\textrm{intent}(\textrm{intent})\prod_i p_{w,i}(w_i)p_{s, i}(s_i)$. To train the SLU model, we minimize the expected value of metric $M$ for each utterance coupled with the cross-entropy loss $CE$ weighted by parameter $\lambda$:
\begin{align}
\theta^* = \mathrm{argmin}_{\theta} \mathrm{E}[M(C)] + \lambda CE
\end{align}
Sub-gradient descent solvers require access to $\nabla_\theta  \mathrm{E}[M(C)]$. In the \emph{sampling approximation} to the term, we use an empirical average of an equivalent quantity,
\begin{align}
\nabla_\theta  \mathrm{E}[M(C)] &= \mathrm{E}[(M(C)-\bar{M}) \nabla_\theta\log p(C; \theta)] \nonumber 
\\ &\approx \frac{1}{n}\sum_{c_i \stackrel{\textrm{iid}}{\sim} p(c;\theta)} (M(c_i) - \bar{M})\nabla_\theta \log p(c_i; \theta),
\end{align} 
where constant $\bar{M}$ is used to reduce the variance of the estimate. 

\noindent In the \emph{n-best approximation},
\begin{align}
\nabla_\theta  \mathrm{E}[M(C)] &\approx \sum_{c\in\bar{\mathcal{C}}} M(c) \nabla_\theta \bar{p}(c; \theta), \\
\bar{p}(c; \theta) &= \frac{p(c;\theta)}{\sum_{c\in\bar{\mathcal{C}}}p(c; \theta)}  \quad \forall c\in\bar{\mathcal{C}}  \label{eq:beam_prob}
\end{align}
where $\bar{\mathcal{C}}$ is a subset of candidates, here the n-best candidates produced by performing \emph{beam-decoding} on the ASR subsystem followed by applying the NLU model to obtain intent, slots for each candidate, is used to obtain a finite-sample approximation of the expectation. Probabilities $\bar{p}(c;\theta)$ are obtained by zeroing out probabilities of candidates not in $\bar{\mathcal{C}}$ and normalizing. 

In either approximation, backpropagation using the non-differentiable metric $M$ is enabled as solvers have access to $\nabla_\theta p(c; \theta)$, as $p(c; \theta)$ is a differentiable function of weights $\theta$. We make use of the n-best approximation in the results section. 

Thus we run both teacher-forcing to obtain $CE_\textrm{total}$ as well as beam-decoding to obtain candidates $\bar{\mathcal{C}}$ as demonstrated in Fig.\ \ref{fig:seq_loss}. As noted in prior work \cite{prabhavalkar2018minimum}, the cross-entropy lends stability to sequence loss training. In Table \ref{tab:seq_loss_different}, we describe the choice of semantic metrics, candidate probability calculations, and regularizing cross-entropy functions for the custom sequence loss training (mSemER, mSLU, mNLU) we propose for joint ASR-NLU model training. We also recover standard mWER training with WER metric, ASR candidate probability and cross-entropy. mSLU-ASR is an example of using semantic sequence losses from an external NLU model for ASR model training. 

\begin{table}
\caption{By varying metric $M$ of interest, candidate probability $p(c; \theta)$, and regularizing $CE$, different sequence loss training methods can be realized for SLU or ASR models. 
\vspace{-0.3cm}
\label{tab:seq_loss_different}}
\begin{tabular}{|p{0.19\linewidth}|p{0.25\linewidth}|p{0.26\linewidth}|p{0.1\linewidth}|}
\hline
\textbf{Training} & \textbf{Metric} $M$ & \textbf{hyp-prob} $p(c; \theta)$ & $CE$ \\ \hline
mWER & WER & \multirow{2}{*}{$\bar{p}(\mathbf{w};\theta)$ (ASR)} & \multirow{2}{*}{$CE_\textrm{asr}$} \\ 
mSLU-ASR & WER + SemER & &  \\ \hline
mSemER & SemER & \multirow{3}{\linewidth}{ $\bar{p}(c; \theta)$ (ASR, NLU) Eq.\ \eqref{eq:beam_prob}} & \multirow{3}{\linewidth}{$CE_\textrm{total}$ as Eq.\ \eqref{eq:x_ent}} \\
mNLU & SemER + IRER + $CE_\textrm{intent}$ &  & \\
mSLU & WER + SemER + IRER + $CE_\textrm{intent}$ &  &  \\ \hline
Transcript-free & SemER + IRER + $CE_\textrm{intent}$ & $\bar{p}(c; \theta)$, Eq.\ \eqref{eq:beam_prob} & $\tilde{CE}_\textrm{total}$, Eq.\ \eqref{eq:transcript_free_xent} \\ \hline   
\end{tabular}
\vspace{-0.7cm}
\end{table}

\vspace{-0.4cm}
\subsection{Application: Transcript-Free Training of ASR models}
\label{sec:transcript_free}
\vspace{-0.2cm}

For ASR model training, ground-truth transcripts are normally required, primarily for the computation of $CE_\textrm{asr}$. We now show how a dataset with audio and only semantic or NLU annotations (intents, slots) and no transcript can be used to update ASR models. This weak label learning problem is motivated by deployments where human transcriptions are not available, but where an inferred semantic feedback from downstream dialogue management systems, applications or user interactions can be used to drive ASR model improvements. We focus on the case where semantic labels are available. In the absence of a reference transcript, the 1-best ASR hypothesis tokens, slots as well as the reference intent $\tilde{c} = \{\tilde{\mathbf{w}}, \tilde{\mathbf{s}}, \textrm{intent}\}$ are treated as the reference in order to prevent catastrophic forgetting of the ASR task. The ASR subsystem is \emph{teacher-forced} \cite{williams1989learning} with the sequence $\tilde{\mathbf{w}}$, and NLU obtains the intent, slots for the resulting sequence. The cross-entropy can be computed as 
\begin{align}
\tilde{CE}_\textrm{total} &= CE_\textrm{intent} -\sum_i \log p_{w, i}(\tilde{w}_i) + \log p_{s, i}(\tilde{s}_i) \label{eq:transcript_free_xent},
\end{align}
without requiring access to a reference transcript. The NLU metrics of ICER, SemER, IRER can be computed from the available labels. The sequence loss training procedure minimizes NLU errors that also results in better ASR performance. 

Note that this is not the only approach to obtaining the cross-entropy regularizer. Teacher ASR or NLU labels or mixing with dataset with transcribed audio are some alternatives.



\begin{table*}[t]
\caption{Performance results on open and proprietary SLU datasets}
\vspace{-0.3cm}
\begin{subtable}{0.44\linewidth}
\vspace{-1.2cm}
\caption{ASR-interface-NLU or Joint modeling approach with mSLU sequence loss training beats all baselines on the test and dev splits of the open Fluent speech dataset on IRER or accuracy \label{tab:fluent}}
\centering
\vspace{-0.2cm}
\begin{tabular}{|p{0.47\linewidth}|p{0.21\linewidth}|p{0.21\linewidth}|}
\hline
{\footnotesize \textbf{Model}} & {\footnotesize Test \textbf{IRER}\%} & {\footnotesize Dev \textbf{IRER}\%}\\ \hline
Transformer audio-intent \cite{radfar2020end} & 2.5 & -\\
Baseline \cite{lugosch2019speech} & 1.2 & -\\
AT-AT (SOTA) \cite{rongali2020exploring} & 0.5 & - \\ \hline
Oracle neural NLU & 0.00 & 0.00 \\
Compositional ASR$\rightarrow$NLU & 0.42 & 2.15 \\
ASR-Gumbel-NLU & 0.40 & 2.05 \\
Joint SLU - no seq training & \textbf{0.39} & 2.05 \\ 
Joint mSLU & \textbf{0.39} & \textbf{1.89} \\ \hline
 \end{tabular}
\end{subtable}
\hspace{0.3 cm}
\begin{subtable}{0.46\linewidth}
\vspace{-0.1cm}
\caption{Comparison of compositional models \& joint models with various sequence loss approaches on the 18-intent eval set of 700k utterances. Performance figures are relative \% improvement from row 2 shown as 0\% \label{tab:crosstown_performance}}
\centering
\vspace{-0.15cm}
\begin{tabular}{|p{0.0175\linewidth}|p{0.31\linewidth}|p{0.13\linewidth}|p{0.15\linewidth}|p{0.13\linewidth}|p{0.13\linewidth}| }
\hline
& \textbf{\footnotesize Model} &{\footnotesize \textbf{WERR}\%} & {\footnotesize \textbf{SemERR}\%}& {\footnotesize \textbf{IRERR}\%}& {\footnotesize \textbf{ICERR}\%}\\ \hline
1&Oracle NNLU & - & 41.17 &  42.93 & 56.34 \\ 
2& Compositional: LAS $\rightarrow$NLU & 0 & 0 & 0 & 0 \\ 
2a& Comp. mWER LAS$\rightarrow$ NLU & 6.23 & 1.07 & 0.96 & 2.82 \\ \hline
3a& LAS-Gumbel-NLU & 2.04 & 1.50 & 0.12 & 3.87 \\ \hline
4a& Joint mSemER & 6.87 & 5.66 & 3.12 & 7.68 \\ 
4b& Joint mNLU & 5.45 & \textbf{5.92} & \textbf{3.26} & 8.91 \\
4c& Joint mSLU & \textbf{7.67} & \textbf{5.91} & 3.08 & \textbf{11.58} \\ \hline
T1 & Transformer-NLU Joint mSemER & 7.46 & \textbf{6.29} & \textbf{4.16} & \textbf{11.62} \\ \hline
\end{tabular}
\end{subtable}
\begin{subtable}{0.44\linewidth}

\vspace{-1.2cm}
\caption{Comparison of compositional baselines and sequence loss approaches on the MoreIntent eval set of 500k utterances. Performance figures are relative \% improvement to row M1 shown as 0\% \label{tab:moreintent_results}}
\centering
\vspace{-0.15cm}
\begin{tabular}{|p{0.05\linewidth}|p{0.31\linewidth}|p{0.13\linewidth}|p{0.16\linewidth}|p{0.14\linewidth}| }
 \hline &\textbf{\footnotesize Model} & {\footnotesize\textbf{WERR}\%} & {\footnotesize \textbf{SemERR}\%}& {\footnotesize\textbf{ICERR}\%}\\ \hline
M1& Compositional & 0 & 0 &  0 \\  
M2& Comp mWER & \textbf{9.40} & 1.59 & 0.71 \\  
M2a& Comp~mSLU-ASR & 6.95 & 3.97 & 0.95 \\ \hline
M3& Joint mSLU & 6.53 & \textbf{6.73} & \textbf{1.98} \\ \hline
\end{tabular}
\end{subtable}
\hspace{0.3 cm}
\begin{subtable}{0.46\linewidth}
\caption{Relative \% improvement from a baseline Joint ASR-NLU model with transcript-free training on the 18 intent and Moreintent datasets \label{tab:weak_nlu_performance}}
\centering
\vspace{-0.2cm}
\begin{tabular}{|p{2.3cm}|c|c|c|}
\hline
{\footnotesize \textbf{Dataset}} & {\footnotesize \textbf{WERR}\%} & {\footnotesize \textbf{SemERR}\%} & {\footnotesize \textbf{ICERR}\%} \\ \hline
18-intent & 2.19 & 5.87 &  10.49 \\ 
Moreintent & 1.01 & 5.88 &  1.77 \\  \hline 
\end{tabular}
\end{subtable}
\vspace{-0.5cm}
\end{table*}

\vspace{-0.4cm}
\section{Data and Experimental Setup}
\label{sec:evaluation}
\vspace{-0.3cm}

We use datasets that include parallel speech transcripts and NLU annotations of intent and slots:
\vspace{-0.2cm}
\begin{itemize}[leftmargin=*]
  \item Fluent speech dataset: Public dataset \cite{lugosch2019speech} of 23k utterances (15 hours) that has been processed to fit the intent, named-entity framework with 10 intents and 2 slots\footnote{Actions are treated as intents. In addition, (inc/dec)rease\_(volume/heat) and (de)activate\_music are added to form 10 intents and 2 slots of object and location.}
  \item 18 intent: Dataset of approximately 5.6M utterances (3.3k hours) with utterances from 18 intents in home automation, global, and notifications  and 40 slots 
  \item More Intent: 22M utterances (16k hours) spanning across 64 intents accounting for 90\% of the data and 122 slots accounting for 99\% of the slots in the data
  \item ASR-only 23k-hour dataset for pretraining the ASR model
\end{itemize}

\noindent\parbox{\linewidth}{\footnotesize  
\noindent\textbf{Training details}: The audio feature is composed of 3 stacked 25 ms LFBE frames with 10 ms shift. This LAS model has 77M parameters: 5x512 BLSTM encoder, 2x1024 LSTM decoder with 4 attention heads of depth 256, projection 728, 4500 subword vocabulary. The NLU model has 4 (text interface)-11 (neural network interface) million parameters with a 2x512 BLSTM encoder, a dense layer for slots, and 2x512 relu feed-forward layers for intent. We also experimented with a 3M parameter Transformer-encoder (2 layers, 8 attention heads, 256 units) NLU model. The LAS model is first pretrained on the 23k hour dataset and finetuned on the specific dataset. With ASR now frozen, NLU is first trained in joint systems followed by joint ASR-NLU fine-tuning using sequence losses. In the 18 intent dataset, NLU is trained in the joint system for 6 epochs followed by sequence loss training  for 2, taking 1 day on 8 Nvidia Tesla V100 GPUs.   
}

\vspace{-0.4cm}
\section{Results and Discussion}
\vspace{-0.3cm}

\noindent\textbf{Sequence loss training beats baselines}

On the open Fluent speech dataset in Table \ref{tab:fluent}, we see all ASR-interface-NLU models beat external baselines that directly extract semantics from audio without intermediate transcript showing utility of ASR pretraining. Both the neural and Gumbel-softmax interface joint models outperform compositional text baselines.The joint model with mSLU sequence loss training is the best-performing model as seen by results on both dev and test splits. This can be categorized as a small dataset of lower semantic complexity as the oracle NLU model perfectly recovers semantics from ground truth transcripts. 

In the 18-intent dataset results of Table \ref{tab:crosstown_performance}, 
the NLU metrics degrade substantially from row 1 (NLU consuming ground-truth transcript) to row 2 (NLU consuming ASR hypothesis), showing impact of ASR errors. In row 2a, the LAS model is further trained with mWER sequence loss, leading to gains in WER as well as NLU metrics. The joint model with mSLU sequence loss training results in best ASR and NLU metrics. From rows 2a and 4b, we see worse WER for the joint model, but better NLU metrics showing that joint training improves ASR performance relevant to downstream NLU. In row T1, mSemER training was used with jointly trained LAS ASR and Transformer-encoder NLU system; this has 1M fewer parameters than joint models with LSTM-based NLU, but shows better performance. 

\noindent\textbf{Sequence loss training optimizes a metric of interest}

Table \ref{tab:crosstown_performance} shows the impact of the non-differentiable metric $M$ to optimize on ASR and NLU performance. mWER training optimizes for WER but this may not reflect its optimal NLU metrics (row 2a vs 4). In rows 4a-c, we use metrics rooted in different definitions of semantic error. The mNLU metric optimizing SemER, IRER, ICER leads to better ICER and IRER than mSemER metric training that optimizes only mSemER. mSLU training (adds WER to mNLU) shows the best ASR performance, reflecting the importance of semantic feedback even for ASR training. Thus we can customize any sequence loss to optimize model performance metric(s). 

\noindent\textbf{Results on a general dataset}

The conclusions from 15 and 3k hours datasets carry over to the large 16k hour \emph{Moreintent} dataset seen in Table \ref{tab:moreintent_results}. Row M2 primarily shows improvements in WER from mWER training of ASR resulting in fewer SLU errors. M2a is an example of semantic sequence loss training of ASR. However, the joint model of M3 trained to optimize SLU metrics shows the best NLU performance. We thus have a recipe to improve ASR and NLU model performance: train an ASR model with mWER sequence loss. The ASR subsystem in the joint model is initialized with these weights and the entire system is trained minimizing SLU sequence losses. 

\noindent\textbf{Both ASR and NLU improve with transcript-free training}

In Table \ref{tab:weak_nlu_performance}, we update models from a common starting point using weak-feedback training with only NLU labels. A 5\% relative improvement in SemER is seen for both the 18-intent and \emph{Moreintent} datasets, and modest ASR improvements suggesting that semantic feedback alone can be used to improve both ASR and SLU. 

\vspace{-0.3cm}
\section{Conclusion}
\vspace{-0.2cm}

Edge deployments of ASR and SLU systems for voice activated assistants require the development of low-footprint performant models. Prior approaches involving either pipelined ASR and NLU models or end-to-end SLU models use the differentiable cross-entropy loss to train, but these do not map to metrics of interest such as word and semantic error rates. In this work, we propose non-differentiable semantic sequence losses and use the REINFORCE framework to train ASR and SLU models. Joint training with custom sequence losses lets ASR be trained with semantic feedback from NLU, and NLU be trained aware of ASR errors. We show that both ASR and NLU performance metrics of SLU systems improve across a range of open and proprietary datasets and beat state-of-the-art models. We also improve and update ASR systems without access to transcripts using weak-feedback via NLU labels alone. 

\vspace{0.2cm}
{\footnotesize \noindent\textbf{Acknowledgement:} We thank Bach, Ehry, Chul, Shehzad, and reviewers for helpful technical comments. Abhinav Khattar assisted with Transformers, Jinxi Guo with mWER discussions, and Zhe Zhang with data preparation. 

\bibliographystyle{IEEEbib}
\footnotesize{\bibliography{seq_loss_slu_arxiv}}

\begin{thebibliography}{10}

\bibitem{graves2012sequence}
Alex Graves,
\newblock ``Sequence transduction with recurrent neural networks,''
\newblock {\em arXiv preprint arXiv:1211.3711}, 2012.

\bibitem{graves2006connectionist}
Alex Graves, Santiago Fern{\'a}ndez, Faustino Gomez, and J{\"u}rgen
  Schmidhuber,
\newblock ``Connectionist temporal classification: labelling unsegmented
  sequence data with recurrent neural networks,''
\newblock in {\em Proceedings of the 23rd international conference on Machine
  learning}, 2006, pp. 369--376.

\bibitem{karita2019comparative}
Shigeki Karita, Nanxin Chen, Tomoki Hayashi, Takaaki Hori, Hirofumi Inaguma,
  Ziyan Jiang, Masao Someki, Nelson Enrique~Yalta Soplin, Ryuichi Yamamoto,
  Xiaofei Wang, et~al.,
\newblock ``A comparative study on {Transformer} vs {RNN} in speech
  applications,''
\newblock in {\em 2019 IEEE Automatic Speech Recognition and Understanding
  Workshop (ASRU)}. IEEE, 2019, pp. 449--456.

\bibitem{chan16}
William Chan, Navdeep Jaitly, Quoc~V. Le, and Oriol Vinyals,
\newblock ``Listen, {Attend} and {Spell}: A neural network for large vocabulary
  conversational speech recognition,''
\newblock in {\em ICASSP}, 2016.

\bibitem{chiu2018state}
Chung-Cheng Chiu, Tara~N Sainath, Yonghui Wu, Rohit Prabhavalkar, Patrick
  Nguyen, Zhifeng Chen, Anjuli Kannan, Ron~J Weiss, Kanishka Rao, Ekaterina
  Gonina, et~al.,
\newblock ``State-of-the-art speech recognition with sequence-to-sequence
  models,''
\newblock in {\em 2018 IEEE International Conference on Acoustics, Speech and
  Signal Processing (ICASSP)}. IEEE, 2018, pp. 4774--4778.

\bibitem{lample2016neural}
Guillaume Lample, Miguel Ballesteros, Sandeep Subramanian, Kazuya Kawakami, and
  Chris Dyer,
\newblock ``Neural architectures for named entity recognition,''
\newblock {\em arXiv preprint arXiv:1603.01360}, 2016.

\bibitem{kim2017onenet}
Young-Bum Kim, Sungjin Lee, and Karl Stratos,
\newblock ``Onenet: Joint domain, intent, slot prediction for spoken language
  understanding,''
\newblock in {\em 2017 IEEE Automatic Speech Recognition and Understanding
  Workshop (ASRU)}. IEEE, 2017, pp. 547--553.

\bibitem{wu2017clinical}
Yonghui Wu, Min Jiang, Jun Xu, Degui Zhi, and Hua Xu,
\newblock ``Clinical named entity recognition using deep learning models,''
\newblock in {\em AMIA Annual Symposium Proceedings}. American Medical
  Informatics Association, 2017, vol. 2017, p. 1812.

\bibitem{chen2019bert}
Qian Chen, Zhu Zhuo, and Wen Wang,
\newblock ``Bert for joint intent classification and slot filling,''
\newblock {\em arXiv preprint arXiv:1902.10909}, 2019.

\bibitem{bunk2020diet}
Tanja Bunk, Daksh Varshneya, Vladimir Vlasov, and Alan Nichol,
\newblock ``Diet: Lightweight language understanding for dialogue systems,''
\newblock {\em arXiv preprint arXiv:2004.09936}, 2020.

\bibitem{hakkani2006beyond}
Dilek Hakkani-T{\"u}r, Fr{\'e}d{\'e}ric B{\'e}chet, Giuseppe Riccardi, and
  Gokhan Tur,
\newblock ``Beyond {ASR} 1-best: Using word confusion networks in spoken
  language understanding,''
\newblock {\em Computer Speech \& Language}, vol. 20, no. 4, pp. 495--514,
  2006.

\bibitem{henderson2012discriminative}
Matthew Henderson, Milica Ga{\v{s}}i{\'c}, Blaise Thomson, Pirros Tsiakoulis,
  Kai Yu, and Steve Young,
\newblock ``Discriminative spoken language understanding using word confusion
  networks,''
\newblock in {\em 2012 IEEE Spoken Language Technology Workshop (SLT)}. IEEE,
  2012, pp. 176--181.

\bibitem{tur2002improving}
Gokhan Tur, Jerry Wright, Allen Gorin, Giuseppe Riccardi, and Dilek
  Hakkani-T{\"u}r,
\newblock ``Improving spoken language understanding using word confusion
  networks,''
\newblock in {\em Seventh International Conference on Spoken Language
  Processing}, 2002.

\bibitem{huang2019adapting}
Chao-Wei Huang and Yun-Nung Chen,
\newblock ``Adapting pretrained transformer to lattices for spoken language
  understanding,''
\newblock in {\em 2019 IEEE Automatic Speech Recognition and Understanding
  Workshop (ASRU)}. IEEE, 2019, pp. 845--852.

\bibitem{haghani2018audio}
Parisa Haghani, Arun Narayanan, Michiel Bacchiani, Galen Chuang, Neeraj Gaur,
  Pedro Moreno, Rohit Prabhavalkar, Zhongdi Qu, and Austin Waters,
\newblock ``From audio to semantics: Approaches to end-to-end spoken language
  understanding,''
\newblock in {\em 2018 IEEE Spoken Language Technology Workshop (SLT)}. IEEE,
  2018, pp. 720--726.

\bibitem{ghannay2018end}
Sahar Ghannay, Antoine Caubri{\`e}re, Yannick Est{\`e}ve, Nathalie Camelin,
  Edwin Simonnet, Antoine Laurent, and Emmanuel Morin,
\newblock ``End-to-end named entity and semantic concept extraction from
  speech,''
\newblock in {\em 2018 IEEE Spoken Language Technology Workshop (SLT)}. IEEE,
  2018, pp. 692--699.

\bibitem{serdyuk2018towards}
Dmitriy Serdyuk, Yongqiang Wang, Christian Fuegen, Anuj Kumar, Baiyang Liu, and
  Yoshua Bengio,
\newblock ``Towards end-to-end spoken language understanding,''
\newblock in {\em 2018 IEEE International Conference on Acoustics, Speech and
  Signal Processing (ICASSP)}. IEEE, 2018, pp. 5754--5758.

\bibitem{qian2017exploring}
Yao Qian, Rutuja Ubale, Vikram Ramanaryanan, Patrick Lange, David
  Suendermann-Oeft, Keelan Evanini, and Eugene Tsuprun,
\newblock ``Exploring {ASR}-free end-to-end modeling to improve spoken language
  understanding in a cloud-based dialog system,''
\newblock in {\em 2017 IEEE Automatic Speech Recognition and Understanding
  Workshop (ASRU)}. IEEE, 2017, pp. 569--576.

\bibitem{tomashenko2020dialogue}
Natalia Tomashenko, Christian Raymond, Antoine Caubri{\`e}re, Renato De~Mori,
  and Yannick Est{\`e}ve,
\newblock ``Dialogue history integration into end-to-end signal-to-concept
  spoken language understanding systems,''
\newblock in {\em ICASSP 2020-2020 IEEE International Conference on Acoustics,
  Speech and Signal Processing (ICASSP)}. IEEE, 2020, pp. 8509--8513.

\bibitem{lugosch2019speech}
Loren Lugosch, Mirco Ravanelli, Patrick Ignoto, Vikrant~Singh Tomar, and Yoshua
  Bengio,
\newblock ``Speech model pre-training for end-to-end spoken language
  understanding,''
\newblock {\em arXiv preprint arXiv:1904.03670}, 2019.

\bibitem{Rao2020Speech}
Milind Rao, Anirudh Raju, Pranav Dheram, Bach Bui, and Ariya Rastrow,
\newblock ``{Speech to Semantics: Improve ASR and NLU Jointly via All-Neural
  Interfaces},''
\newblock in {\em Proc. Interspeech}, 2020, pp. 876--880.

\bibitem{williams1992simple}
Ronald~J Williams,
\newblock ``Simple statistical gradient-following algorithms for connectionist
  reinforcement learning,''
\newblock {\em Machine learning}, vol. 8, no. 3-4, pp. 229--256, 1992.

\bibitem{juang1997minimum}
Biing-Hwang Juang, Wu~Hou, and Chin-Hui Lee,
\newblock ``Minimum classification error rate methods for speech recognition,''
\newblock {\em IEEE Transactions on Speech and Audio processing}, vol. 5, no.
  3, pp. 257--265, 1997.

\bibitem{prabhavalkar2018minimum}
Rohit Prabhavalkar, Tara~N Sainath, Yonghui Wu, Patrick Nguyen, Zhifeng Chen,
  Chung-Cheng Chiu, and Anjuli Kannan,
\newblock ``Minimum word error rate training for attention-based
  sequence-to-sequence models,''
\newblock in {\em 2018 IEEE International Conference on Acoustics, Speech and
  Signal Processing (ICASSP)}. IEEE, 2018, pp. 4839--4843.

\bibitem{guo2020efficient}
Jinxi Guo, Gautam Tiwari, Jasha Droppo, Maarten~Van Segbroeck, Che-Wei Huang,
  Andreas Stolcke, and Roland Maas,
\newblock ``{Efficient Minimum Word Error Rate Training of RNN-Transducer for
  End-to-End Speech Recognition},''
\newblock in {\em Proc. Interspeech}, 2020, pp. 2807--2811.

\bibitem{keneshloo2019deep}
Yaser Keneshloo, Tian Shi, Naren Ramakrishnan, and Chandan~K Reddy,
\newblock ``Deep reinforcement learning for sequence-to-sequence models,''
\newblock {\em IEEE transactions on neural networks and learning systems}, vol.
  31, no. 7, pp. 2469--2489, 2019.

\bibitem{jang2016categorical}
Eric Jang, Shixiang Gu, and Ben Poole,
\newblock ``Categorical reparameterization with {Gumbel}-softmax,''
\newblock {\em arXiv preprint arXiv:1611.01144}, 2016.

\bibitem{williams1989learning}
Ronald~J Williams and David Zipser,
\newblock ``A learning algorithm for continually running fully recurrent neural
  networks,''
\newblock {\em Neural computation}, vol. 1, no. 2, pp. 270--280, 1989.

\bibitem{radfar2020end}
Martin Radfar, Athanasios Mouchtaris, and Siegfried Kunzmann,
\newblock ``{End-to-End Neural Transformer Based Spoken Language
  Understanding},''
\newblock in {\em Proc. Interspeech}, 2020, pp. 866--870.

\bibitem{rongali2020exploring}
Subendhu Rongali, Beiye Liu, Liwei Cai, Konstantine Arkoudas, Chengwei Su, and
  Wael Hamza,
\newblock ``Exploring transfer learning for end-to-end spoken language
  understanding,''
\newblock {\em arXiv preprint arXiv:2012.08549}, 2020.

\end{thebibliography}

\end{document}